\documentclass[runningheads]{llncs}

\usepackage[T1]{fontenc}
\usepackage{graphicx,verbatim, subcaption}
\usepackage{amssymb}
\usepackage{xcolor}
\usepackage{array}
\usepackage{caption}
\usepackage{amsmath}

\begin{document}
\title{Bias and Generalizability of Foundation Models across Datasets in Breast Mammography}

\author{Elodie Germani\inst{1} \and
Ilayda Selin-Türk\inst{2} \and Fatima Zeineddine\inst{3} \and Charbel Mourad\inst{3} \and
Shadi Albarqouni\inst{1,2,4}}
\authorrunning{E. Germani et al.}
\titlerunning{Bias \& Generalizability of FMs for Mammography}

\institute{Clinic for Diagnostic and Interventional Radiology, University Hospital Bonn, Bonn, Germany  \and TUM School of Computation, Information and Technology, Technical University of Munich, Munich, Germany \and Department of Diagnostic Imaging and Interventional Therapeutics, Lebanese Hospital Geitaoui, Beyrouth, Lebanon \and Helmholtz AI, Helmholtz Munich, Munich, Germany \\ \email{shadi.albarqouni@ukbonn.de}}
 
\maketitle   

\begin{abstract}
Over the past decades, computer-aided diagnosis tools for breast cancer have been developed to enhance screening procedures, yet their clinical adoption remains challenged by data variability and inherent biases. Although foundation models (FMs) have recently demonstrated impressive generalizability and transfer learning capabilities by leveraging vast and diverse datasets, their performance can be undermined by spurious correlations that arise from variations in image quality, labeling uncertainty, and sensitive patient attributes. In this work, we explore the fairness and bias of FMs for breast mammography classification by leveraging a large pool of datasets from diverse sources—including data from underrepresented regions and an in-house dataset. Our extensive experiments show that while modality-specific pre-training of FMs enhances performance, classifiers trained on features from individual datasets fail to generalize across domains. Aggregating datasets improves overall performance, yet does not fully mitigate biases, leading to significant disparities across under-represented subgroups such as extreme breast densities and age groups. Furthermore, while domain-adaptation strategies can reduce these disparities, they often incur a performance trade-off. In contrast, fairness-aware techniques yield more stable and equitable performance across subgroups. These findings underscore the necessity of incorporating rigorous fairness evaluations and mitigation strategies into FM-based models to foster inclusive and generalizable AI.


\end{abstract}

\section{Introduction}
Breast cancer is one of the most significant global health challenges, with over 2.3 million new cases and approximately 670,000 deaths reported in 2022 alone~\cite{bray2024global}. Early and accurate detection is crucial for improving patient outcomes, and mammographic screening, typically confirmed by biopsy, remains a cornerstone of clinical diagnosis. In recent years, deep learning models have shown promise in aiding radiologists by extracting breast cancer biomarkers with high performance, sometimes even surpassing that of human experts~\cite{mckinney2020international}. However, these models are often developed using datasets drawn predominantly from specific populations, which tend to under-represent marginalized groups, potentially leading to biases and reliance on spurious correlations that do not generalize well across populations~\cite{ricci_lara_addressing_2022}. This under-representation is particularly problematic in breast cancer detection, as critical risk factors such as age and breast density may vary across different ethnicities, and geographic regions~\cite{checka_relationship_2012,del_carmen_mammographic_2007,heller_breast_2015}.

In response to these challenges, foundation models (FMs) have emerged as a promising solution due to their ability to learn rich and transferable visual representations from diverse large-scale datasets~\cite{Bommasani2021FoundationModels,paschali_foundation_2025}. By working on pre-extracted features rather than raw images, FMs offer the potential for improved generalizability and reduced computational overhead in resource-limited settings~\cite{girdhar_imagebind_2023}. However, recent studies have revealed that FMs are also susceptible to bias, as they can inadvertently capture spurious correlations inherent in their training data~\cite{glocker_risk_2023,jin2024fairmedfm}. Such biases raise concerns about the equity of AI systems in clinical practice, particularly when deployed across diverse demographic groups.

Motivated by these observations, this work investigates the presence of bias in FMs applied to breast cancer biomarkers detection and explores bias mitigation strategies through domain adaptation and fairness techniques. Unlike previous works primarily assessing FM fairness within individual datasets~\cite{jin2024fairmedfm}, we extend our analysis to between-dataset biases and domain shifts. To this end, we aggregate a diverse set of mammography datasets sourced from various parts of the world, including under-represented regions, and supplement them with an in-house dataset from Lebanon (LBMD) with around 3,000 images from 700 patients. Directly sourced from clinical practice, LBMD captures real-world complexities often overlooked in curated public datasets, offering an additional perspective on clinical settings. Our \textbf{contributions} are threefold. First, we conduct a comprehensive analysis of bias in FMs by evaluating the risk of spurious correlations when classifiers are trained on different datasets. Second, we assess traditional domain-adaptation and fairness strategies as potential solutions to mitigate these biases. Third, by incorporating the LBMD dataset, we demonstrate the clinical relevance of our results, addressing disparities in breast cancer biomarkers detection, and ultimately advancing the development of more robust and equitable AI tools to support radiologists in diverse clinical settings.

\section{Methodology}
\label{sec:method}

Let $\mathcal{X}$ be the space of mammography images and $\mathcal{Y}$ the label space (\textit{e.g.} $\{0,1,2\}$ for diagnosis or $\{1,2,3,4\}$ for breast density classification). Each data point is a triplet $(x_i, y_i, d_i)$, where $x_i \in \mathcal{X}$ is the image, $y_i \in \mathcal{Y}$ its label, and $d_i \in \mathcal{D}$ denotes the domain or dataset source. The complete dataset is given by \(\mathcal{S} = \{(x_i, y_i, d_i)\}_{i=1}^N\). Our goal is to learn a classifier $f_\theta: \mathcal{X} \to \mathcal{Y}$ that achieves high predictive performance while mitigating any sort of bias.

\paragraph{\textbf{Feature Extraction and Classification.}}
We extract a representation $\phi(x) \in \mathbb{R}^{m}$ using a frozen, pre-trained FM $\phi$. A linear probe is then trained over these features: \(
f_\theta(\phi(x)) = \text{sigmoid}(W\,\phi(x) + b),
\) where $\theta = \{W, b\}$.
We use a weighted cross-entropy loss: \(
\mathcal{L}_{\text{WCE}}(\theta; \mathcal{S}) = \frac{1}{N} \sum_{i=1}^{N} w_{y_i} \cdot \ell (f_\theta(\phi(x_i)), y_i)\), with $w_{y_i}$ corresponding to inverse class percentages. Simple minimization of $\mathcal{L_{\text{WCE}}}$ may correct for class imbalance, but might not adequately address domain shift or bias. 

\paragraph{\textbf{Bias and Domain-Adaptation Methods.}}
We investigate several strategies to promote domain invariance and mitigate biases. Specifically, we explored two categories of techniques to balance classifiers' performance across datasets: i) domain-adaptation strategies (\verb|DANN|, \verb|FairDisCO|, and \verb|FADES|) primarily designed to learn domain-invariant representations, and ii) fairness-aware strategies (\verb|GroupDRO|, \verb|MOE|) which explicitly focus on reducing performance disparities across subgroups. Table~\ref{tab:methods} summarizes the formulations of each method. \\

\begin{table}[t!]
\centering
\caption[Summary of Bias Mitigation and Domain Adaptation Methods.]{Summary of bias mitigation and domain-adaptation Methods. }
\label{tab:methods}
\scriptsize
\begin{tabular}{p{2cm} p{10cm}}
\hline
\textbf{Method} & \textbf{Objective} (Note: $z_i = \phi_z(\phi(x_i))$, and $H(\cdot|\cdot)$ is conditional entropy) \\
\hline
\verb|DANN|~\cite{ganin2016domain} & {$\mathcal{L}_{\mathrm{DANN}}(\theta;\psi) = \mathcal{L_{\mathrm{WCE}}} -\ell_d\Bigl(g_\psi(\phi(x_i)), d_i\Bigr)$} \\
\hline
\verb|FairDisCO|~\cite{du2022fairdisco} & {$\mathcal{L}_{\mathrm{FairDisCO}}(\theta; \psi; \phi_{z}) =  \mathcal{L_{\mathrm{DANN}}} + \alpha \mathcal{L}_{\mathrm{conf}} + \beta \mathcal{L}_{\mathrm{contr}}$} \newline  $\mathcal{L}_{\mathrm{conf}}(\theta) = -\sum_{i=1}^{N} \frac{1}{N} \cdot \text{log}(f_\theta(\phi(x_i)))$  \newline $
\mathcal{L}_{\mathrm{contr}}(\theta, \phi_{z}) = \sum_{(i,j) \in P_y} \log \frac{\exp(\text{sim}(z_i, 
z_j))}{\exp(\text{sim}(z_i, z_j))) + \sum_{k \in N_y} \exp(\text{sim}(z_i, z_k)))}$ \newline 
$\mathcal{L}_{conf}$ max. equal probabilities across $\mathcal{D}$, $\mathcal{L}_{contr}$ max. $\mathcal{D}$-invariant representations.
 \\
\hline
\verb|FADES|~\cite{jang2024fades} & {$\mathcal{L}_{\mathrm{FADES}}(\theta; \psi; \phi_z) = \mathcal{L_{\mathrm{DANN}}} +\mathcal{L}_{\mathrm{TC}} + \mathcal{L}_{\mathrm{CMI}} + \mathcal{L}_{\mathrm{reg}}$}
 \newline 
$\mathcal{L}_{\mathrm{CMI}}(\theta, \psi) = I_{\phi}(f_\theta(z_i); g_\psi(\phi(z_i)) | d_i)$ and $\mathcal{L}_{\mathrm{TC}}(\phi_z) = D_{\text{KL}}(z_i \parallel \prod_{j} z_j)$ aim to learn disentangled representations: domain-relevant, task-relevant and irrelevant.
\newline $\mathcal{L}_{\mathrm{reg}} = - \left( H(f_\theta(z_i) | z_R) + H(g_\psi(z_i) | z_R) \right)$ to regularize training objective.\\
\hline
\verb|GroupDRO|~\cite{sagawa2019distributionally} & {$\mathcal{L}_{\mathrm{GroupDRO}} = \min_\theta \max_{q\in \Delta^{|\mathcal{D}|}} \sum_{d\in \mathcal{D}} q_d\,\mathcal{L}_d(\theta)$} \newline \scriptsize $\mathcal{L}_d(\theta) = \frac{1}{|\mathcal{S}_d|}\sum_{i:d_i=d}\ell\Bigl(f_\theta(\phi(x_i)), y_i\Bigr)$ to minimize empirical worst-group risk.\\
\hline
\verb|MOE|~\cite{li2023sparse} & 
{$\mathcal{L}_{\mathrm{MOE}}(\theta)=\frac{1}{N}\sum_{i=1}^N \ell\Bigl(f_{\mathrm{MOE}}(x_i), y_i\Bigr)$}, with each expert specialized in one domain. \newline $f_{\mathrm{MOE}}(x)=\sum_{e=1}^E \alpha_e(x) f^e_\theta(\phi(x))$ and $\alpha_e(x) = \frac{\exp(w_e^T x)}{\sum_{j=1}^{E} \exp(w_j^T x)}
$\\
\hline
\end{tabular}
\end{table}

\noindent \textit{Domain Adversarial Neural Network}(\verb|DANN|) 
uses adversarial learning to enforce extraction of domain-invariant features from \(\phi_z(\cdot)\) by introducing a domain classifier $g_\psi(\cdot)$ and reversing the gradient of the domain classification loss 
$\ell_d$~\cite{ganin2016domain}.

\noindent \textit{Fair Disentanglement with Contrastive Learning} (\verb|FairDisCO|) employs adversarial and contrastive learning. It encourages samples from different domains with the same label to be close in a new feature space \(\phi_z(\cdot)\) to mitigate bias~\cite{du2022fairdisco}.

\noindent \textit{Fair Disentanglement with Sensitive Relevance} (\verb|FADES|) penalizes \(\phi_z(\cdot)\) features predictive of domain while maintaining those correlated with both domain and target tasks. It integrates total correlation (TC), conditional mutual information (CMI), and adversarial loss to minimize domain information leakage~\cite{jang2024fades}.

\noindent \textit{Group Distributionally Robust Optimization} (\verb|GroupDRO|) explicitly optimizes for the worst-case performance across domains. By re-weighting the loss based on each domain’s \(\mathcal{D}\) performance, \verb|GroupDRO| ensures that the model does not favor majority groups at the expense of under-represented ones~\cite{sagawa2019distributionally}.

\noindent \textit{Mixture-of-Experts} (\verb|MOE|) uses a set of expert classifiers \(f^e_\theta(\cdot)\), each specializing in different domains, and combines their outputs through a gating mechanism \(\alpha_e(x)\). This allows the model to adaptively leverage domain-specific expertise while benefiting from a shared representation, as described in~\cite{li2023sparse}.
\section{Experiments and Results}
\paragraph{\textbf{Foundation Models.}}
We consider several FMs drawn from recent reviews~\cite{paschali_foundation_2025,moor_foundation_2023}. \textbf{MammoCLIP}~\cite{ghosh_mammo-clip_2024} was trained on 25,355 mammograms from the UPMC dataset using contrastive multi-view learning and yields 2,048-dimensional features via its EN-B5 encoder. In contrast, \textbf{MedCLIP}~\cite{wang_medclip_2022} and \textbf{GLORIA}~\cite{Huang_2021_ICCV} were developed on 500,000 and 200,000 X-ray images respectively, both employing a ResNet-50 backbone to produce 512-dimensional embeddings. Additionally, \textbf{CLIP}~\cite{radford_learning_2021} was trained on 400 million internet-sourced image-text pairs with contrastive learning, while \textbf{DINOv2}~\cite{oquab_dinov2_2023} uses a self-distillation framework on 142 million images to generate lightweight representations of size 384. 

\begin{table}[t!]
    \caption{Overview of mammography datasets with available scans after selection and splitting. The number of samples in the training sets is shown in parentheses.}
    \centering
    \resizebox{\textwidth}{!}{
    \begin{footnotesize}
    \begin{tabular}{ccccccccccc}
        \hline
          & \textbf{CBIS-DDSM} & \textbf{RSNA} & \textbf{INbreast} &  \textbf{MIAS} & \textbf{CMMD} & \textbf{VinDR} & \textbf{CDD-CESM} & \textbf{KAU-BCMD} & \textbf{MMD} & \textbf{LBMD} \\
         \hline
         Country & US & US/AU & Portugal & UK & China & Vietnam & Egypt & KSA & Iraq & XXX \\ 
         Sites & 4 & 2 & 1 &  & 2 & 2 & 1 & 1 & 4 & 1\\         
         \hline
         Patients & 1,391 & 1,970 & 115 & 165 & 1,277 & 930 & 326 & 442 & 745 & 696 \\
         Scans & 2,844 & 9,594 & 410 & 322 & 2,742 & 3,709 & 1,003 & 1,774 & 745 & 3,090 \\ 
         Age (y) & \textit{N/A} & 59 ± 11 & \textit{N/A} & \textit{N/A} & 47 ± 11 & 44 ± 12 & 50 ± 12 & 49 ± 7 & \textit{N/A} & 58 ± 11 \\
         \hline 
         Diagnosis & $\checkmark$ & $\checkmark$ & - & $\checkmark$ & $\checkmark$ & -& $\checkmark$ & - & $\checkmark$ & $\checkmark$ \\
         Benign & 1,253 (875) & 1,487 (1,039) &  & 64 (47) & 1,102 (774) & & 331 (252) &  & 0 & 1,993 (1,392) \\
         Malignant & 1,220 (860) & 1,069 (734) &  & 52 (31) & 1,640 (860) & & 331 (239) &  & 125 (88) & 2 (2) \\
         \hline 
         Density & $\checkmark$ & $\checkmark$ & $\checkmark$ & $\checkmark$ & - & $\checkmark$ & $\checkmark$ & $\checkmark$ & - & $\checkmark$ \\
        A & 396 (273) & 529 (377) & 136 (97) & 106 (82) &  & 12 (12) & 8 (8) & 577 (399) &  & 375 (231) \\
        B & 1,103 (760) & 2,789 (1,904) & 146 (108) & 104 (70) &  & 337 (226) & 329 (247) & 827 (600) &  & 1,050 (746) \\ 
        C & 879 (633) & 2,861 (2,095) & 99 (64) & 112 (77) &  & 2,852 (1,968) & 515 (315) & 332 (208) &  & 1,012 (717) \\ 
        D & 464 (325) & 343 (209) & 28 (17) & 0 &  & 508 (390) & 70 (62) & 108 (80) &  & 180 (135) \\   \hline
    \end{tabular}
    \end{footnotesize}
    }
    \label{tab:datasets}
\end{table}

\paragraph{\textbf{Datasets.}} We use mammography datasets from diverse countries and institutions, ensuring a representative analysis. Our collection includes four prominent datasets; the Digital Database for Screening Mammography \textbf{(CBIS-DDSM)}~\cite{cbisddsm} from the USA, the RSNA Screening Mammography Breast Cancer Detection Dataset \textbf{(RSNA)}~\cite{rsna} from the USA and Australia, \textbf{INbreast}~\cite{moreira_inbreast_2012} from Portugal, and Mammographic Image Analysis Society \textbf{(MIAS)}~\cite{suckling_mammographic_2015} from the UK. To further capture diversity and address the under-representation of certain regions, we integrated datasets including the Chinese Mammography Database \textbf{(CMMD)}~\cite{cai_online_2023}, \textbf{VinDr-Mammo}~\cite{nguyen_vindr-mammo_2023} from Vietnam, the Categorized Digital Database for Low Energy and Subtracted Contrast Enhanced Spectral Mammography images \textbf{(CDD-CESM)}~\cite{khaled_categorized_2022} from Egypt, the King Abdulaziz University Breast Cancer Mammogram Dataset \textbf{(KAU-BCMD)}~\cite{alsolami_king_2021} from Saudi Arabia, and the Mammogram Mastery dataset \textbf{(MMD)}~\cite{aqdar_mammogram_2024} from Iraq. Additionally, we incorporated the Lebanese Breast Mammography Dataset \textbf{(LBMD)}, an internally curated collection co-developed with our clinical partners at the Lebanese Hospital Geitaoui and assembled exclusively for this project; all cases within the \textbf{LBMD} are biopsy-confirmed and the data collection protocol received full ethical approval for use in this work.

\paragraph{\textbf{Sample selection.}} Our combination of datasets was initially highly imbalanced, with some datasets containing over 50,000 samples (\textit{e.g.} RSNA) while others had as few as 300 (\textit{e.g.} MIAS). Additionally, the class imbalance was significant within datasets; for instance, 75\% of VinDR samples belong to density class C. To minimize these imbalances and focus on dataset biases, we applied a sample selection strategy to have more balanced classes. First, we dropped samples with no labels for diagnosis or density class, \textit{i.e.} the two classification tasks investigated. All labels were aggregated from the original metadata, where benign and malignancy classes were biopsy-confirmed in most datasets. We categorized patients into three diagnosis classes: healthy, benign, and malignant. We capped each class at 1,000 patients, randomly sampling when necessary while retaining all available patients in smaller classes. For VinDR, as this dataset did not contain the diagnosis information, we applied our sampling selection strategy at the density level. Finally, datasets were split at the patient level into training (70\%) and test (30\%) sets, ensuring no data leakage. Table~\ref{tab:datasets} provides more details on the composition of each dataset after the sample selection strategy. 

\paragraph{\textbf{Implementation details.}} Images were preprocessed using the framework proposed by~\cite{ghosh_mammo-clip_2024}. We used a rule-based approach to crop images according to the breast ROI. We set values less than 40 to 0 and eliminated consistently identical rows and columns, supposing these denote background. The final images had a size of 1,520×912. Experiments were implemented in Python v3.10 using Pytorch v2.4.1. {Individual} classifiers were trained on each dataset, and \verb|Unified| on the aggregated datasets for two tasks: diagnosis and breast density. We searched for optimal batch size (8, 16, or 32) and learning rate (1e-3, 1e-4, or 1e-5) using 3-fold cross-validation within the training set. Hyperparameters giving the best accuracy after 20 epochs were then used for training on the whole set for 50 epochs. We used the same hyperparameter optimization for all mitigation strategies, except for \verb|FADES|. Due to its computational cost, we fixed the batch size to 32, the learning rate to 1e-4, and trained for 30 epochs. Technical details specific to each strategy are reported in the code and will be publicly available upon acceptance. 
We computed differences between F1 score distributions across datasets using one-sided Wilcoxon test for statistical significance.
To evaluate the classifiers' fairness, we computed Equal Opportunity Difference (EOD) and Average Odds Difference (AOD) across subgroups \(g_1, g_2 \in \mathcal{G}\) and labels \(k \in \mathcal{Y}\):
$\text{EOD} = \left(P(\hat{Y} = k \mid Y = k, G = g_1) - P(\hat{Y} = k \mid Y = k, G = g_2)\right)$, and $\text{AOD} = \text{EOD} + (P(\hat{Y} = k \mid Y \neq k, G = g_1) - P(\hat{Y} = k \mid Y \neq k, G = g_2)$.

\subsection{Exploring the feature embeddings}
\begin{figure}[t!]
    \centering
    \includegraphics[width=\linewidth]{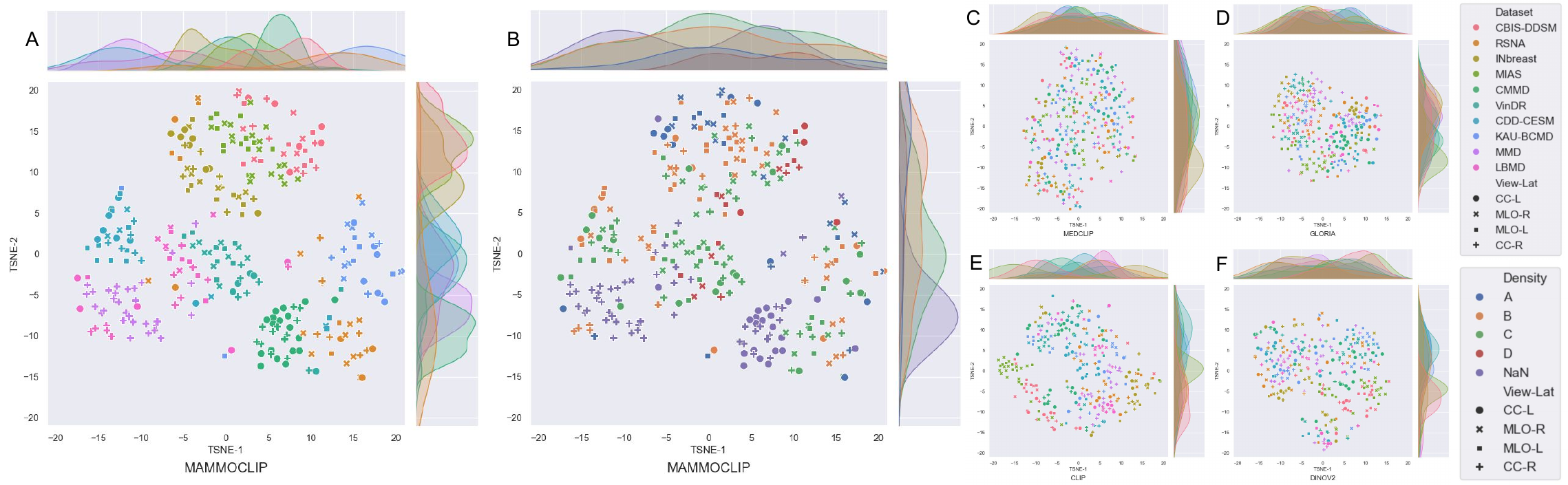}
    \caption{
    t-SNE visualization of MammoCLIP, color-coded by dataset (\textbf{A}) and density (\textbf{B}), and FMs: GLORIA (\textbf{C}), MedCLIP (\textbf{D}), CLIP (\textbf{E}), DINOv2 (\textbf{F}).}
    \label{fig:features}
\end{figure}

Fig.~\ref{fig:features} presents a t-SNE-based analysis of features extracted from FMs. The results illustrate distinct clustering behaviors, reflecting how each FM encodes mammography-specific characteristics. There is no visible clustering according to view, lining with MammoCLIP's multi-view learning strategy. For other FMs, this suggests their ability to learn view-invariant features, likely through data augmentation. MammoCLIP exhibits well-defined clusters, with features from the same dataset tightly grouped, suggesting strong dataset-specific encoding. Smoother patterns emerge for CLIP and DINOv2, where at least one t-SNE component captures dataset-specific information. Features from MedCLIP and GLORIA are widely dispersed, with no clear dataset-specific clustering. These models seem to learn more generalized feature representations, likely due to pre-training on diverse medical images. Interestingly, breast density attributes seem to impact feature distributions along the t-SNE components, with a smooth transition from low- to high-dense breasts in MammoCLIP features.

\subsection{Robustness of classifiers to domain-shift} 
Tab.~\ref{tab:f1score} shows the performance of classifiers trained using features extracted from each FM for diagnosis and breast density classification. For both tasks, \verb|Unified| classifiers trained on features from MammoCLIP outperform those based on GLORIA ($p<0.05$, average relative improvement of +15.3\%), CLIP ($p<0.01$, +9.7\%), and DinoV2 ($p<0.01$, +13.3\%), highlighting the advantages of pre-training on modality-specific data compared to domain-related (\textit{i.e.} X-rays) or natural images. It is worth mentioning that MedCLIP-based classifiers exhibit notably poor performance across all tasks, suggesting that the extracted features may be predominantly noisy. Overall, classifiers have high performance when tested on the same dataset they were trained on (\textit{Indiv. (internal)}), with average F1 scores of 0.73 and 0.53 for MammoCLIP on diagnosis and breast density, respectively. However, when tested on other datasets (\textit{Indiv. (external)}), a drastic drop in F1 scores (up to $-50\%$ for diagnosis) was remarkable, potentially due to overfitting of dataset-specific characteristics encoded in MammoCLIP's features. 

\newcolumntype{C}[1]{>{\centering\let\newline\\\arraybackslash\hspace{0pt}}m{#1}}

\begin{table}[t!]
    \caption{F1 score of classifiers: mean ± standard deviation across datasets, overall best performance is in \textbf{bold}. Stars indicate significantly different performance: {\color{red}red} stars for lower than MammoCLIP, {\color{green}green} stars for higher than Unified, {\color{blue}blue} stars for lower than Unified. \scriptsize * = $p<0.01$, ** = $p<0.05$.}
    \label{tab:f1score}
    \centering
    \scriptsize
    \begin{tabular}{p{2cm}|C{1.9cm}|C{1.9cm}|C{1.9cm}|C{1.9cm}|C{1.9cm}}
         \multicolumn{6}{c}{\textbf{Diagnosis}} \\
         \hline
         & MammoCLIP & MedCLIP & GLORIA & CLIP & DinoV2 \\
         \hline
         Indiv. (internal) & 0.73 ± 0.11 & 0.37 ± 0.21 & 0.68 ± 0.13 & 0.61 ± 0.14 & 0.62 ± 0.12 \\
        Indiv. (external) & 0.32 ± 0.11 & 0.18 ± 0.07 & 0.28 ± 0.08 & 0.24 ± 0.10 & 0.32 ± 0.10 \\
        \hline
        Indiv. (overall) & 0.37 ± 0.22\textsuperscript{\color{blue}*} & 0.25 ± 0.18\textsuperscript{\color{blue}*} & 0.38 ± 0.19\textsuperscript{\color{blue}*} & 0.34 ± 0.2\textsuperscript{\color{blue}*} & 0.39 ± 0.19\textsuperscript{\color{blue}*} \\
        \verb|Unified| & \textbf{0.65 ± 0.14} & 0.32 ± 0.27\textsuperscript{\color{red}**} & 0.58 ± 0.13\textsuperscript{\color{red}*} & 0.56 ± 0.14\textsuperscript{\color{red}**} & 0.57 ± 0.16\textsuperscript{\color{red}**} \\
        \hline
        \verb|DANN|~\cite{ganin2016domain} & 0.54 ± 0.14 & 0.18 ± 0.16 & 0.42 ± 0.14 & 0.46 ± 0.11 & 0.47 ± 0.07 \\
        \verb|FairDisCO|~\cite{du2022fairdisco} & 0.63 ± 0.14 & 0.18 ± 0.16 & 0.43 ± 0.14 & 0.56 ± 0.15 & 0.54 ± 0.18 \\
        \verb|FADES|~\cite{jang2024fades} & 0.62 ± 0.16 & 0.39 ± 0.17 & 0.51 ± 0.17 & 0.54 ± 0.19 & 0.56 ± 0.15 \\
        \hline
        \verb|MOE|~\cite{li2023sparse} & 0.64 ± 0.13 & 0.38 ± 0.04 & 0.43 ± 0.13 & 0.49 ± 0.15 & 0.55 ± 0.16 \\
        \verb|GroupDRO|~\cite{sagawa2019distributionally} & 0.57 ± 0.14 & 0.32 ± 0.27 & 0.49 ± 0.15 & 0.47 ± 0.16 & 0.54 ± 0.16 \\ 
        \hline 
        \multicolumn{6}{c}{\textbf{Density}} \\
        \hline
         & MammoCLIP & MedCLIP & GLORIA & CLIP & DinoV2 \\
         \hline
        Indiv. (internal) & 0.53 ± 0.19 & 0.32 ± 0.18 & 0.41 ± 0.19 & 0.41 ± 0.19 & 0.50 ± 0.13 \\
        Indiv. (external) & 0.42 ± 0.08 & 0.2 ± 0.04 & 0.31 ± 0.10 & 0.33 ± 0.12 & 0.4 ± 0.07 \\
        \hline
        Indiv. (overall) & 0.41 ± 0.16\textsuperscript{\color{blue}*} & 0.22 ± 0.15\textsuperscript{\color{blue}*} & 0.32 ± 0.16\textsuperscript{\color{blue}*} & 0.33 ± 0.16\textsuperscript{\color{blue}*} & 0.4 ± 0.14\textsuperscript{\color{blue}*} \\
        \verb|Unified| & 0.59 ± 0.17 & 0.19 ± 0.08\textsuperscript{\color{red}**} & 0.5 ± 0.15\textsuperscript{\color{red}*} & 0.5 ± 0.16\textsuperscript{\color{red}**} & 0.51 ± 0.15\textsuperscript{\color{red}**} \\
        \hline
        \verb|DANN|~\cite{ganin2016domain} & 0.56 ± 0.16 & 0.19 ± 0.08 & 0.47 ± 0.14 & 0.5 ± 0.14 & 0.48 ± 0.11 \\
        \verb|FairDisCO|~\cite{du2022fairdisco} & 0.54 ± 0.17 & 0.19 ± 0.08 & 0.47 ± 0.14 & 0.49 ± 0.14 & 0.48 ± 0.12 \\
        \verb|FADES|~\cite{jang2024fades} & 0.57 ± 0.16 & 0.19 ± 0.08 & 0.49 ± 0.15 & 0.54 ± 0.16\textsuperscript{\color{green}*} & 0.51 ± 0.15 \\
        \hline
        \verb|MOE|~\cite{li2023sparse} & 0.55 ± 0.17 & 0.19 ± 0.08 & 0.42 ± 0.13 & 0.46 ± 0.13 & 0.5 ± 0.16 \\
        \verb|GroupDRO|~\cite{sagawa2019distributionally} & \textbf{0.66 ± 0.08\textsuperscript{\color{green}*}} & 0.19 ± 0.08 & {0.52 ± 0.11} & {0.57 ± 0.06} & {0.56 ± 0.08} \\
        \hline
    \end{tabular}

\end{table}

\subsection{Effectiveness of bias and domain-adaptation strategies} 
\verb|Unified| classifiers show performance similar to individual classifiers on their test sets (internal), with F1 score variations of ±15\%. Aggregating datasets effectively improves generalization compared to individual classifiers (overall F1 score improved by $+75\%$ and $+49\%$, $p<0.05$ with \verb|Unified| vs. Indiv.). However, \verb|Unified| classifiers exhibit performance disparities, with F1 score standard deviations of 0.15 across test datasets, suggesting that such aggregation cannot fully mitigate biases. \verb|DANN| shows slightly lower overall F1 scores than \verb|Unified| classifiers, especially for the diagnosis task. While this technique aims to learn domain-invariant representations, it seems to do so at the cost of overall performance. Similar observations can be made for \verb|FairDisCO|, \verb|FADES|, and \verb|MOE|. For Breast density, \verb|GroupDRO| produces consistently tighter F1 score distributions than \verb|Unified| classifiers with standard deviations of 0.08, 0.06, and 0.08 for MammoCLIP, CLIP, and DinoV2, respectively, indicating reductions of disparities across test datasets. Additionally, it outperforms the \verb|Unified| classifier across FMs, \textit{e.g.} MammoCLIP with a relative improvement of 12\% ($p<0.01$). For diagnosis, mitigation strategies do not improve overall performance nor reduce disparities compared to \verb|Unified|, likely due to variability in diagnostic label availability across datasets, \textit{e.g.} no benign samples for MMD.

\subsection{Bias and domain-adaptation in under-represented subgroups}

\begin{figure}[t!]
    \includegraphics[width=\linewidth]{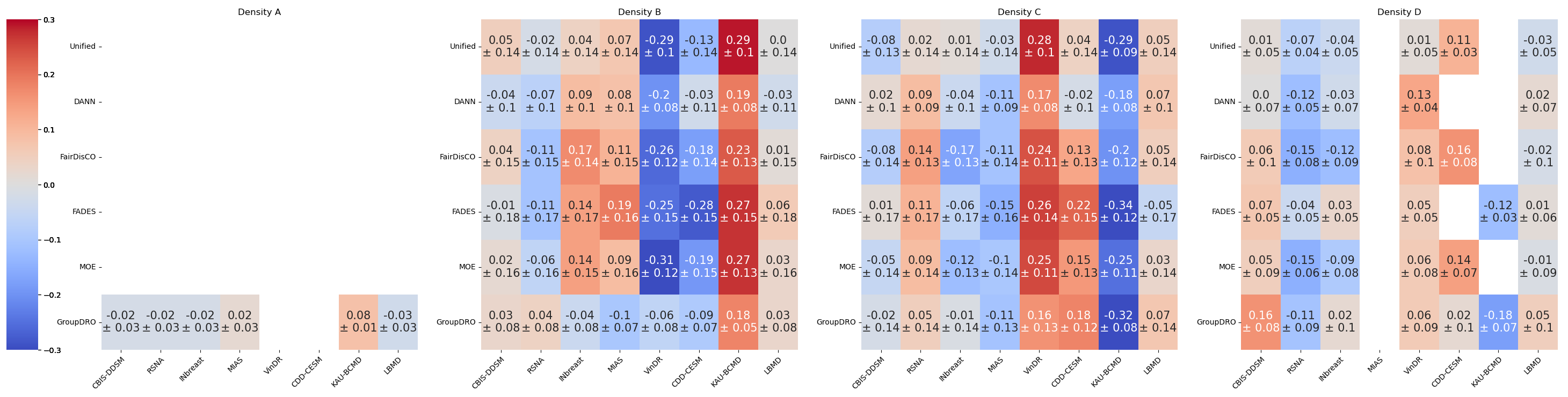}
    \centering \includegraphics[width=0.8\linewidth]{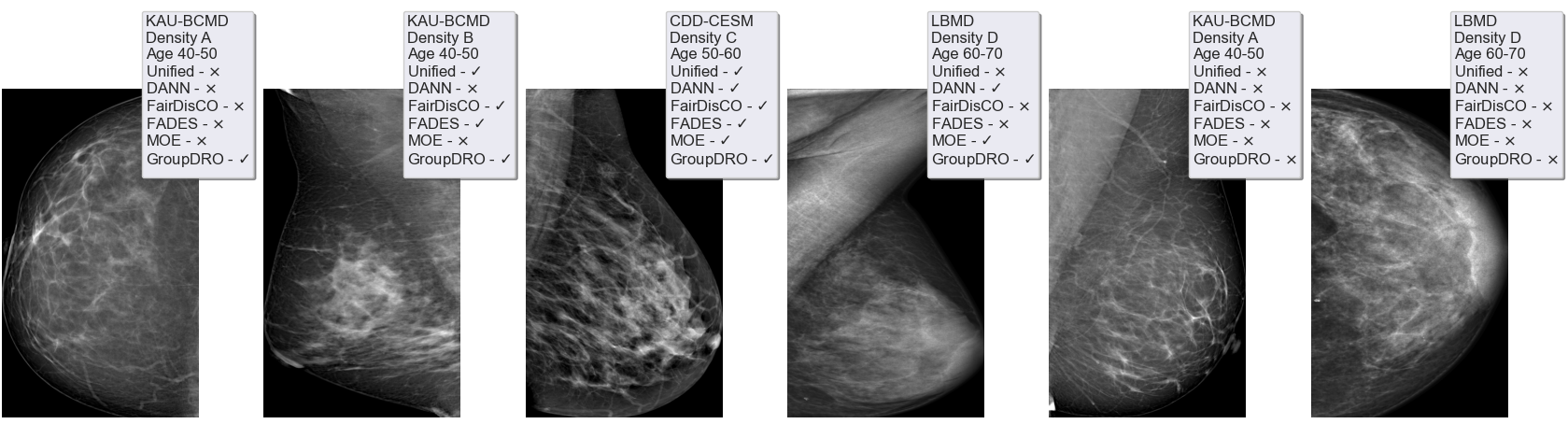}
    \caption{(\textit{Top}) AOD scores across datasets for MammoCLIP on breast density. \newline
    (\textit{Bottom}) From left to right: samples from different density groups from A to D, and from under-represented subgroups (density A, age < 40 and density D, age >70)). $\checkmark$ indicates correctness, $\times$ represents misclassification.}
    \label{fig:fairness-breakdown}
\end{figure}

Certain subgroups are under-represented in our datasets, \textit{e.g.} breast density classes A and D representing 11 and 9\% of the dataset and age <40 (10\%) and >70 (10\%), and are unequally represented across datasets (see Tab.~\ref{tab:datasets}). Fig.\ref{fig:fairness-breakdown} (top) illustrates prediction disparities across datasets and breast density classes, where $AOD\simeq0$ indicates fair performance. \verb|GroupDRO| and \verb|DANN| achieve the most fair performance for all breast density classes ($AOD_{max} \simeq 0.2$, $AOD_{min}\simeq-0.3$ and $AOD_{avg.}\simeq0$ across breast density classes and datasets), aligning with \verb|DANN|’s domain-invariant feature learning strategy. However, for \verb|DANN|, this fairness comes at the cost of performance (see Tab.~\ref{tab:f1score}). \verb|GroupDRO| stabilizes performance across breast density classes, notably improving prediction for class A, which other classifiers struggled with. This ability to learn across domains, while favoring under-represented subgroups, is critical for extreme breast densities (A, D) and age groups due to their strong interplay~\cite{checka_relationship_2012}. Fig.~\ref{fig:fairness-breakdown} (bottom) presents samples from different subgroups and classifiers’ successes and failures in breast density classification. Variations in contrast, texture, and patterns across classes and datasets may introduce spurious correlations, underscoring the need for fairness-aware strategies. \verb|GroupDRO| seems an effective strategy to mitigate these biases and could be further refined to incorporate more fine-grained attributes.
\section*{Conclusion}

This paper explores biases in FM for breast mammography classification. Our analysis reveals that modality-specific pre-training of FM is beneficial for performance, but individual classifiers still fail to generalize well beyond their training data. Aggregating datasets enhances overall performance, emphasizing the need for broader dataset contributions. However, this strategy is not sufficient to mitigate biases, leading to disparities across under-represented subgroups. Domain-adaptation strategies address these disparities, but often at the cost of performance. On the other hand, fairness-aware techniques ensure more equitable performance across under-represented subgroups and classes. These findings have significant implications for deploying AI-driven mammography analysis in clinical practice. FMs must be equipped with fairness-aware optimization techniques to limit the risk of reinforcing existing biases. Future work should investigate fairness-aware FMs, in addition to federated learning frameworks, fostering more inclusive and generalizable medical AI solutions.

\section*{Acknowledgments}
This article was made possible through partial support from the German Academic Exchange Service (DAAD) with funds from the Federal Foreign Office (AA). It was also developed within the interdisciplinary framework of the Arab-German Young Academy of Sciences and Humanities (AGYA), which is funded by the German Federal Ministry of Education and Research (BMBF) under grant 01DL20003. The authors thank their colleagues at the Lebanese Hospital Geitaoui, Beyrouth, Lebanon for their support. 

\bibliographystyle{splncs04}
\bibliography{references}

\end{document}